# Large Language Model-Powered Conversational Agent Delivering Problem-Solving Therapy (PST) for Family Caregivers: Enhancing Empathy and Therapeutic Alliance Using In-Context Learning

**Liying Wang, Ph.D.[1], Daffodil Carrington, M.S.[2], Daniil Filienko, M.S.[3], Caroline El Jazmi, M.S.[4], Serena Jinchen Xie, M.S.[5], Martine De Cock, Ph.D.[3], Sarah Iribarren, Ph.D.[6], Weichao Yuwen, Ph.D.[7]**

[1]College of Nursing, Florida State University, Tallahassee, FL, USA; [2]School of Nursing, University of Washington, Seattle, WA, USA; [3]School of Engineering and Technology, University of Washington, Tacoma, WA, USA; [4]Department of Computer Science, University of Texas at Austin, Austin, TX, USA; [5]School of Medicine, University of Washington, Seattle, WA, USA; [6]Department of Biobehavioral Nursing and Health Informatics, University of Washington, Seattle, WA, USA; [7]School of Nursing & Healthcare Leadership, University of Washington Tacoma, Tacoma, WA, USA

**Abstract**

*Family caregivers often face substantial mental health challenges due to their multifaceted roles and limited resources. This study explored the potential of a large language model (LLM)-powered conversational agent to deliver evidence-based mental health support for caregivers, specifically Problem-Solving Therapy (PST) integrated with Motivational Interviewing (MI) and Behavioral Chain Analysis (BCA). A within-subject experiment was conducted with 28 caregivers interacting with four LLM configurations to evaluate empathy and therapeutic alliance. The best-performing models incorporated Few-Shot and Retrieval-Augmented Generation (RAG) prompting techniques, alongside clinician-curated examples. The models showed improved contextual understanding and personalized support, as reflected by qualitative responses and quantitative ratings on perceived empathy and therapeutic alliances. Participants valued the model's ability to validate emotions, explore unexpressed feelings, and provide actionable strategies. However, balancing thorough assessment with efficient advice delivery remains a challenge. This work highlights the potential of LLMs in delivering empathetic and tailored support for family caregivers.*

**Introduction**

Caregiving has increasingly become a public concern as the aging population continues to grow. Family caregivers in the US reached 53 million in 2024[1]. Almost one in six non-caregivers are estimated to start providing caregiving services within 2 years [2]. Informal or unpaid family caregivers make up the majority of the caregiving workforce, assisting in household tasks, providing personal care, emotional support, and managing medical care[3]. About half of caregivers have provided care for at least two years and almost one third do so for at least 20 hrs every week [2]. Despite the amount and intensity of care, about half of family caregivers reported not receiving any type of support (e.g., respite care, financial support, mental health counseling)[1]. The accumulating stress and lack of coping resources led to adverse health outcomes among family caregivers, including having 14 or more unhealthy days over the past months (15% for mentally unhealthy and 18% for physically unhealthy), insufficient sleep (37%), having two or more chronic diseases (41%) [1,3]. During the COVID-19 pandemic, 57.2% of caregivers screened positive for symptoms of an anxiety or depressive disorder which was three times the prevalence compared to non-caregivers [4].

While emotional and psychological needs are evident among family caregivers [5–7], obtaining mental health support is challenging. Mental health professionals are in shortage in the US and around the world. In the US, there are only 30 clinical psychologists and 16.6 psychiatrists for every 100,000 people, and in rural areas, these numbers are lower[8]. More than one-third of the US population live in mental health professional shortage areas[9]. Family caregivers, in addition to the waiting time and high cost, face additional challenges including: logistical challenges in managing personal work schedule, caregiving schedule, and other family responsibilities such as parenting; low priority of self-care in the face of competing demands; feeling guilty or shame when taking time for oneself [10–12].

Large language model-based chatbots expanded the possibilities of delivering mental health support, compared to previous rule-based and learning-based chatbots[13]. LLMs demonstrated utilities in mental disorder diagnosis, depression management, suicide risk evaluation, and providing accessible mental health support [14,15]. Empathy is a key clinical element that strengthens therapeutic relationship[16]. LLMs exhibit the core elements of cognitive empathy, including emotion recognition and providing emotionally validating responses aligned with the contexts [17]. LLMs also demonstrated higher emotional awareness and provided more empathetic responses compared to human benchmarks [18,19]. However, the majority of LLM-based chatbots developed for caregivers focused on developing information support for family caregivers [20–22]. One LLM-based chatbot was fine-tuned with Problem-Solving Therapy (PST) session scripts and was tested on a group of family caregivers, but the study was more focused on the sentiment of chatbot responses than its ability to effectively deliver mental health support [23].

Digital health interventions designed for caregivers show effectiveness in increasing caregiver coping skills, emotion regulation, motivation to self-care, and decreasing stress and burden.[24] Most interventions draw from Cognitive Behavioral Therapy (CBT), an evidence-based therapy for multiple mental health conditions. However, given its emphasis on cognitive strategies that take time and practice, CBT may be particularly challenging for caregivers who have competing demands and limited cognitive bandwidth. [25–27] For example, the cognitive restructuring exercises in CBT, while valuable, may feel like an additional burden rather than immediate relief for caregivers dealing with practical day-to-day challenges. In contrast, Problem-Solving Therapy (PST) offers a more action-oriented approach that aligns better with caregivers' needs [28]. PST's focus on identifying concrete problems and developing practical solutions makes it more accessible and immediately applicable to caregivers' situations [29]. The skills learned through PST are also easier to generalize across different caregiving challenges, providing caregivers with a flexible toolkit that can be adapted as their responsibilities evolve.

PST has shown effectiveness in many mental health interventions, including for depression, anxiety disorder, PTSD, and substance use [30,31]. In addition, PST interventions were found effective in releasing caregiver burden and enhancing coping, self-efficacy, depression, and quality of life among family caregivers [32,33]. Our work on designing tools for caregivers began with a rule-based PST therapy chatbot (Caring of Caregivers Online, COCO)[34], which would give a response from a set of human-curated options based on the user's demonstrated need [35]. Our most recent work extended rule-based COCO to utilize LLMs to deliver PST[36]. The step-by-step approach of PST can steer LLMs to generate structured guidance, while the flexibility of LLMs allows for the incorporation of other therapeutic frameworks alongside PST. Filienko et al.[36] highlighted two key gaps in COCO's performance: the tendency to express empathy by resorting to repetitive phrases and the limited exploration of symptom contexts. Thus, the authors experimented with incorporating additional skills to the LLM-based therapy chatbot to enhance empathy and improve context exploration.

To address these challenges, we propose to integrate Motivational Interviewing (MI) and behavioral chain analysis (BCA) to enhance empathic response and contextual exploration of COCO. Motivational Interviewing (MI), involves using collaborative techniques to enhance engagement and strengthen personal motivation and commitment to a specific goal [37,38]. MI can be integrated with PST to deepen empathy and validation while affirming individuals strengths to move from contemplation to action to reach their goal. BCA involves examining the sequence of events, thoughts, feelings, and actions that lead to a problematic behavior [39]. Applying BCA to LLM prompting can guide the model to better consider the broader context and emotional nuances of a caregiver's situation. This study aims to continue to enhance COCO's contextual understanding of caregivers before providing advice, which would contribute to perceived empathy, therapeutic alliance, and tailored advice-giving.

**Methods**
*Study Design*
We used a convergent mixed-methods approach, collecting quantitative and qualitative evaluation data [40]. A within-subject design with randomized counterbalancing was employed, where each participant interacted with four LLMs and completed an evaluation survey on perceived empathy and therapeutic alliance from the interaction.

*Participants and procedure*
Participants were recruited using the crowdsourcing platform Prolific with the following eligibility criteria: 18 years or older, residing in the US, fluent in English, self-reported as a family caregiver, and having a Prolific approval rating of at least 95%. Individuals were excluded if they had participated in prior studies on the same LLM-powered chatbot. Eligible participants were provided with a link to the study's web-based user interface (UI). Each participant was asked to describe their caregiving experiences, then interacted with four LLM models for 5 minutes and completed at least 8 turns before completing the evaluation survey. The four models were randomly sequenced.

After completing the conversations and evaluations of all four models, participants were provided with a completion code and redirected to Prolific to submit their study.

*Study measures*
Therapeutic alliance is defined as a collaborative and trusting relationship between a client and a therapist that enables effective communication and goal-setting during the therapeutic process. This was measured using the Session Rating Scale, a 4-item scale designed to measure four dimensions of therapeutic alliance, including relationship (to what extent you felt heard, understood, and respected by the chatbot), goals & topics (to what extent the session focused on what you wanted to work on and talk about), approach or method (to what extent the therapy approach is a good fit for you), and overall satisfaction (overall rating of interaction) [41]. Participants responded on a 0-10 scale, with higher scores indicating better model performance on that dimension. A total score was calculated by summing up ratings on each dimension to indicate an overall model performance on therapeutic alliance.

Empathy was measured by a 3-item scale, with each item measuring a dimension of empathy, including emotional reactions (to what extent you felt warmth and care from the chatbot), interpretations (to what extent the chatbot accurately reflected your feelings and experiences back to you), and explorations (to what extent the chatbot helped you explore feelings you hadn't expressed) [42]. Participants responded on a 0-2 Likert scale, where a higher score indicated better model performance. A total score was calculated by summing up all three items to indicate overall model performance on empathy. Additionally, we conducted automatic evaluation of empathy, which uses a base RoBERTa classifier to quantify the model's emotional reactions, interpretations, and explorations from model responses to users [42].

We collected qualitative data through open-ended questions tailored to participants' ratings to gather detailed feedback on model performance. For instance, participants who rated the emotional reaction dimension of empathy as 0 (out of 2) were asked to elaborate on what made the chatbot feel cold or uncaring, while those who rated it 2 (out of 2) were asked to share specific actions that made them feel particularly cared for. Ratings between 0 and 2 prompted a request for general feedback. Similarly, ratings below 3 on any SRS item were followed by questions seeking specific examples of suboptimal performance, ratings between 3 and 8 prompted general feedback, and ratings above 8 were followed by prompts for examples of positive model performance.

*Model development*
We created 9 distinct LLM models, each utilizing different combinations of in-context learning techniques. An internal study was conducted to identify the top-performing models to help mitigate the effects of evaluator fatigue. The four models selected are listed below:
- Model 1 (PST with Few Shot, GPT-4o): The baseline model is the best-performing model configuration from the study by Filienko et al. [36]. This model maintained the validated PST structure with zero-shot prompting and curated examples demonstrating core PST principles for family caregivers.
- Model 2 (PST and MI Integration with Few-shot, GPT-4o): Enhances the baseline by integrating MI directives into both the base prompt and few-shot examples. The base prompt structure emphasizes change talk and partnership strengthening while incorporating therapeutic techniques such as affirmation, reflection, and autonomy emphasis. The few-shot examples range from single-turn demonstrations to multi-turn dialogues, designed to enhance the model's capability to conduct coherent and contextually relevant conversations within the therapeutic framework.
- Model 3 (PST, MI, and BCA integration using RAG, GPT-4o): Builds upon Model 2's MI-enhanced base prompt and few-shot examples, adding BCA instructions through zero-shot prompting and implementing RAG architecture. The dual-component system comprises a transformer-based retriever (HuggingFace sentence-transformers) and GPT-4o generator, enabling dynamic BCA instruction retrieval and real-time semantic matching for contextual adaptation [43].
- Model 4 (PST, MI, and BCA integration using RAG, Llama 3): With the rise of the quality of open-source models, we tested the Llama 3 70 billion parameter model with MI Few shot and chain analysis prompts similar to Model 3. This family of models, being comparable to GPT-4 in a variety of tasks and small enough to be runnable on a commercial-level GPU demonstrates a promising new direction for future applications. They permit the developers to produce quality answers without relying on external services or disclosing the data to third parties, hence providing a chance for both high-quality and privacy-conscious text generation, especially important for marginalized communities.

*Analysis*

Participant conversations were examined to ensure conversations aligned with the study purpose. We conducted descriptive analyses and repeated-measures Analysis of Variance (ANOVA) on quantitative evaluation data to compare model performance on empathy and therapeutic alliance. Qualitative data from open-ended questions were analyzed using a complementary approach, systematically examining participants' narrative responses to validate, expand, or clarify the quantitative findings. This approach allowed us to triangulate findings across methods, using qualitative data to provide context and inform interpretation to the quantitative metrics [40]. Additionally, we conducted ad hoc analysis to compare the word count of the responses from different models and examine associations between word count and evaluations of model performance, after observing that some models' outputs were longer than others.

**Results**

*Participants*

In total, 37 participants were recruited from Prolific, 33 completed the study, and 28 were included in data analysis. Five participants were excluded from data analysis as they did not follow the instructions on conversing with the chatbot (e.g., directly and repetitively asking for resources without presenting a caregiving challenge first). The study included 8 male participants and 20 female participants ($N$=28). Participants' ages ranged from 21-55 years old with an average age of 33 years old. Twelve participants reported being White (42.86%), 11 Black (39.29%), 4 Mixed (14.29%), and 1 Other (3.57%). Caregiving experience ranged anywhere from 6 months to over 10 years.

*Empathy*

Based on the sum score from human evaluation of all three dimensions, Model 4 was the best-performing model overall ($M$=5.21, $SD$=1.07). According to the descriptive statistics comparing model performance results on the three empathy dimensions, Model 4 was the top performing model in interpretation ($M$=1.89, $SD$=0.31; how well the model understands users' feelings and experience) and exploration dimensions ($M$=1.77, $SD$=0.51; how well the model helps the participants explore unexpressed feelings). Model 2 was rated the highest among all four models in the emotional reaction dimension ($M$=1.79, $SD$=0.42; how well the model expressed warmth and care towards you). The omnibus repeated-measure ANOVA on human evaluation results suggested no statistically significant differences for interpretation ($F(3, 81)$=1.333, $p$=0.269) and the total empathy score ($F(3, 81)$=0.941, $p$=0.425). The omnibus tests were not conducted for emotional reaction and exploration due to insufficient variance in data. Pairwise comparisons found no statistically significant differences between model pairs across all dimensions.

Based on the algorithm-rated scores across the three empathy dimensions, there were notable patterns in model performance. For emotional reaction, Model 2 received the highest mean score ($M$=0.93, $SD$=0.60). For interpretation, all models performed similarly, with no significant differences observed across versions. For exploration, Model 3 was rated the highest ($M$=1.24, $SD$=0.98). The omnibus repeated-measures ANOVA results revealed statistically significant differences for emotional reaction ($F(3, 81) = 3.12$, $p$=0.031) and explorations ($F(3, 81)$=6.70, $p$<0.001). However, no statistically significant differences were found for Interpretations ($F(3, 81)$=0.42, $p$=0.742) (Table 1). Post-hoc pairwise comparisons indicated that for Emotional Reactions, significant differences were observed between Model 2 and Model 3 ($t(27)$=3.27, $p$=0.049). For Explorations, significant differences were found between Model 3 and Model 4 ($t(27)$=4.11, $p$=0.001). No significant pairwise differences were found for Interpretations, suggesting similar performance across models for this dimension.

Table 1: Empathy rating from human and algorithm evaluations in *Mean* (*SD*) ($N$=28)

| Model | Emotional Reactions | | Interpretations | | Explorations | | Total Score | |
|---|---|---|---|---|---|---|---|---|
| | Human | Algorithm | Human | Algorithm | Human | Algorithm | Human | Algorithm |
| 1 | 1.57 (0.57) | 0.81 (0.63) | 1.68 (0.55) | 0.00 (0.11) | 1.54 (0.69) | 1.02 (1.00) | 4.79 (1.57) | 0.81 (0.63) |
| 2 | 1.79 (0.42)* | 0.93 (0.60)* | 1.75 (.052) | 0.01 (0.12)* | 1.46 (0.74) | 1.07 (1.00) | 5.00 (1.49) | 0.93 (0.60)* |
| 3 | 1.64 (0.49) | 0.78 (0.65) | 1.71 (0.46) | 0.01 (0.16)* | 1.46 (0.79) | 1.24 (0.98)* | 4.82 (1.31) | 0.78 (0.65) |
| 4 | 1.74 (0.45) | 0.81 (0.61) | 1.89 (0.31)* | 0.01 (0.11)* | 1.77 (0.51)* | 0.74 (0.96) | 5.21 (1.07)* | 0.81 (0.61) |

* indicated the highest score in a given dimension.

The qualitative data also support the strong performance of Model 4 in understanding their feelings and experiences. Participants shared how the model recognized nuanced emotions, validated their perspectives, and offered tailored solutions. For example, one participant wrote: "When it replied regarding caregiving and personal lives being intertwined and how that can make someone always feel "on the clock" I felt understood." Participants also praised Model 4's ability to guide deeper reflection on unexpressed emotions and provide insights that helped them uncover underlying feelings. Although some participants found the approach of model 4 slightly clinical, many appreciated the thoughtful questions and emphasis on fostering self-awareness and boundary setting. For example, one participant wrote: "They (the model) asked very thoughtful questions and otherwise reframed things in a way that made me realize what I hadn't considered." For Model 2, participants' feedback on the emotional reaction dimension reflected mixed experiences, where some felt care and attentiveness from the chatbot, while others perceived the replies to be more rapid than they preferred. Positive feedback from participants showed that they felt understood and that the chatbot conveyed concern toward them.

Participants generally left less negative or constructive feedback during model evaluation. Overall, model 2 was perceived as not showing enough warmth and lacking depth in its understanding (e.g., It felt less caring at times, like a bot, not a person; Sometimes it missed the nuances of what I was feeling and just agreed without offering much insight). Model 3 faced limitations in emotion exploration, where participants commented that: It avoided going deeper when I mentioned sensitive topics, which I felt like a missed opportunity. While model 4 received generally favorable feedback, some participants found the conversation a little "scripted" and lacked "genuine curiosity."

*Therapeutic Alliance*
The total score across all therapeutic alliance dimensions suggested that Model 4 was the top-performing model overall ($M$=32.51, $SD$=7.01). Descriptive statistics for the individual dimensions highlighted differences in model performance. Model 4 was rated highest in three dimensions: goals and topics ($M$=8.36, $SD$=1.77; how well the model addressed relevant topics and goals), overall satisfaction ($M$=8.04, $SD$=1.82; participants' overall satisfaction with the interaction), and relationship ($M$=8.15, $SD$= 2.01; how well the model built rapport with participants). Model 3 was rated the highest for approach or method ($M$=8.04, $SD$=1.90; how well the model's approach aligned with the participants' expectations).

We counted the number of times each model received the highest rating among the four models for each dimension of the session rating scale, as well as the overall total across all dimensions. In cases where multiple models shared the highest rating for a given dimension (e.g., both models received a 10), all models with the same top rating were counted as 'rated the highest.' Consistent with the mean scores presented in Table 2, Model 4 excelled in 'the Goals and Topics' and 'Relationship' dimensions, while Model 3 was most frequently ranked at the top for 'Overall Satisfaction' and the total ratings across dimensions (Figure 1).

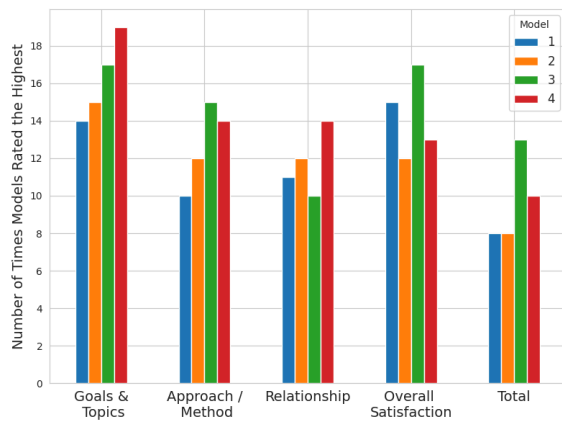

**Figure 1.** Top-Rated Models Across Different Evaluation Dimensions of Therapeutic Alliance

The omnibus repeated-measure ANOVA results indicated no significant differences among the models for any dimension of therapeutic alliance: relationship ($F$(3, 81)=0.357, $p$=0.784), goal and topics ($F$(3, 81)=1.766, $p$=0.158), approach or method ($F$(3, 81)=1.020, $p$=0.387), overall satisfaction ($F$(3, 81)=0.218, $p$=0.884), and total score ($F$(3, 81)=0.757, $p$=0.521) (Table 2). Pairwise comparisons similarly showed no statistically significant difference between model pairs for any dimension. For Goals and Topics, model 1 versus model 4 showed the strongest trend toward significance ($t(27) = -1.752$, $p=0.086$).

Table 2: SRS rating from human and algorithm evaluations in Mean (SD) (N=28)

| Model | Session Rating Scale (0-10); M (SD) | | | | |
|---|---|---|---|---|---|
| | Relationship | Goals and Topics | Approach or Method | Overall Satisfaction | Total Score |
| | Human | Human | Human | Human | Human |
| 1 | 7.61 (2.385) | 7.29 (2.46) | 7.14 (2.40) | 7.61 (2.54) | 29.64 (9.12) |
| 2 | 8.07 (2.19) | 8.29 (1.94) | 7.71 (1.96) | 7.86 (2.26) | 31.93 (7.66) |
| 3 | 7.86 (1.94) | 8.25 (1.60) | 8.04 (1.90)* | 8.04 (2.08) | 32.18 (7.04) |
| 4 | 8.15 (2.01)* | 8.36 (1.77)* | 7.96 (2.13) | 8.04 (1.82)* | 32.51 (7.01)* |

* indicated the highest score in a given dimension.

The qualitative responses supported the findings above. Participants generally felt respected and understood by Model 4. The model's non-judgmental stance and thoughtful responses helped build rapport, creating a positive interaction. For example, participants wrote: "It was not at all judgy about coping methods I've used, which made me feel safe; The AI wanted to understand and made thoughtful efforts to clarify my feelings." Participants also found model 4 to be highly responsive to participants' needs, maintained focus on goals and topics, provided tailored responses, and stayed aligned with participants' priorities (e.g., "This session felt particularly focused and relevant to my needs because the chatbot zeroed in on the specific challenges I'm facing as a caregiver—balancing my own self-care with the demands of caregiving and the guilt that comes with it. The questions it asked helped me unpack my emotions and identify the root causes of my guilt and anxiety, and the advice was tailored to my situation. Instead of offering generic advice, it offered practical, actionable steps that directly addressed my current struggles, like incorporating small self-care moments and asking for small favors from others.").

Participants appreciated the overall positive and constructive nature of their interactions with Model 4. Many found the conversions with Model 4 valuable and uplifting. For example, one participant wrote: "Overall, the session felt like a meaningful step toward balancing self-care with caregiving in a realistic, manageable way, which is exactly what I need right now." Participants found Model 3's approach aligned with their expectations and that the strategies offered were thoughtful and clear. They also appreciate the overall structure and the adaptability of the model through the interaction, which allowed them to more effectively explore solutions. They specifically pointed out the use of follow-up questions that brought in alternative perspectives and enhanced session depth.

Participants pointed out a few limitations of the models as well. Model 1 was perceived as having a narrower focus in terms of goals and topics, where participants commented, "I felt it ignored some parts of my problem." Model 2 was perceived as a little "rigid" in following a structure and some participants found it difficult to direct the conversations to obtain suggestions within the limited turns. Similarly, participants found Model 4 respectful, empathetic, and helpful in guiding them to break down large tasks into smaller steps, but they also expressed disappointment in not receiving the suggestions that they were expecting. Model 3 received mixed feedback where some participants found it lacking in a nuanced understanding of the "peculiarities of my caregiving experience", while others reported that the model gave personalized suggestions that took into consideration their caregiving responsibilities.

*Relationship between Word Count And Clinical Metrics*
The words per turn ranged from 66.63 (Model 2, $SD=22.6$) to 133.15 (Model 4, $SD=39.62$). Repeated measures ANOVA revealed significant differences in the number of words per turn across four models ($F(3, 81)=26.96$, $p<0.001$). Pairwise t-tests showed that model 4 consistently produced significantly longer responses compared to model 1 ($t(27)= -4.236$, $p=0.001$) and model 2 ($t(27)=-11.092$, $p<0.001$).

Pearson's Correlation did not find significant correlations between words per turn and clinical metrics (empathy and therapeutic alliance). For empathy, weak positive correlations were found with emotional reaction ($r=0.161$, $p=0.094$)) and total empathy score ($r=0.171$, $p=0.075$), although these were not statistically significant. Similarly, no significant correlations were observed between words per turn and therapeutic alliance.

**Discussion**
This is one of the first studies that developed an LLM-powered chatbot to deliver evidence-based mental health intervention for family caregivers. We used a combination of in-context techniques (Few-Shot learning and RAG) to implement the prompts written by clinicians (a clinical psychologist and nurses) on the team. Among the four models evaluated by caregivers, Model 4 (all techniques combined using Llama 3) received the most favorable ratings on empathy and session rating scales, although not statistically significant. Our ad hoc analysis revealed that Model 4 also tended to generate longer responses compared to the rest of the models. These findings pointed to a combination of prompting techniques and clinician-developed prompts as an effective approach to continue to optimize the performance of LLMs in providing empathetic mental health support.

The ratings on empathy and therapeutic alliance both pointed to Model 4 as the best-performing model. Although all models were perceived as empathetic, model 4 and model 3, the two models that utilized BCA, received positive comments on the thorough assessment, reflecting the benefits of implementing BCA. However, the drawback of this approach was that some participants were not able to receive advice within the first eight turns in this study context. BCA and clinical assessment, in general, are time-consuming but essential to providing tailored suggestions. Related, some participants expressed a preference to receive advice directly rather than being prompted to think about effective strategies they may have used. This reflected that some participants may prefer efficiency in addressing their needs and find it less effortful to receive advice rather than engage in collaborative generation of solutions. Indeed, effectively taking into account client preference in therapeutic approaches benefits therapeutic alliance [44]. This indicates that future iterations of the LLM-based chatbot may consider user preference and experiment with balancing assessment to collect enough information and provide practical advice to users to address their immediate needs. Other alternatives may include providing a rationale for conducting a thorough assessment and offering options for users to choose from with varied emphasis on assessment vs. advice-giving.

Ratings of model performance on empathy and therapeutic alliance lacked variance, which limits our ability to differentiate model performance using statistical tests. Additionally, participants generally provided more positive feedback than negative or constructive feedback. Positive comments were typically more detailed and thorough, whereas negative or constructive feedback was often brief and less elaborate. This is potentially due to the source of participants and data. While utilizing crowdsourcing platforms, like Prolific, can have its benefits, such as increased diversity, it also has its limitations [45]. Crowdsourced participants are less likely to fully participate in online evaluations compared to traditional, in-person tests due to the lack of a moderator who can correct mistakes in real-time [46]. Participants are also paid for their time and participation, which can negatively impact the quality and thoughtfulness of participant responses and feedback, as high rewards are not necessarily correlated with high test quality [46]. Future studies should consider using a combination of approaches, such as real-time user interviews and surveys to obtain more constructive feedback.

**Limitations**
There are several limitations of this study. First, the conversion length was constrained to 8 turns and 5 minutes and the majority of participants stopped at this minimum requirement. This limited duration, combined with the use of chain analysis techniques, may have insufficiently captured the chatbot's full capabilities. In several interactions, the model might not be able to complete the assessment and progress to provide suggestions due to limited turns. Second, while the average study completion time was 80 minutes (interaction and evaluation time combined), previous studies indicated that test performance quality typically deteriorates after 40 minutes due to participant fatigue and diminishing engagement [47]. Third, in the current evaluation protocol, we did not require evaluators to interact with all models using the same scenario. As a result, we observed participants discussing different scenarios with different models, leading to a lack of standardization of scenarios across models. This introduced variability in the level of complexity and characteristics of scenarios, which might have acted as confounding variables when assessing the models' performances. Future studies could address this by restricting the heterogeneity of scenarios, for example, by providing a curated pool of common caregiving challenges of similar characteristics for participants to use to interact with models. Additionally, there is a lack of dialogue corpora or automated metrics specifically for evaluating empathy in multi-turn conversations and therapeutic alliances [48]. The current study relied solely on self-reported ratings for therapeutic alliance, which are susceptible to individual biases or expectations regarding chatbots, AI, or therapy sessions. Additionally, the automated empathy algorithm used was not trained on multi-turn conversations but rather on single-turn post-response format data, limiting its applicability to assessing multi-turn chatbot conversations [42]. Future research could explore the development and adoption of automated metrics to measure constructs related to therapeutic alliance and empathy. Lastly, this study focused on feasibility and preliminary effects of LLM-powered chatbot delivered PST. Future studies with larger sample size are needed to examine the generalizability and reliability of these findings.

**Conclusion**
This study demonstrated the effectiveness of using a combination of prompting techniques to enhance LLM performance in delivering mental health support for family caregivers. Specifically, the integration of Few-Shot prompting with clinician-curated examples and RAG allowed for more dynamic and contextually appropriate responses, while the incorporation of BCA and MI enhanced assessment depth and personalization of suggestions. The in-context learning techniques and clinical techniques enabled the models to better emulate domain-specific therapeutic skills, with the best-performing model showing strong capabilities in empathy and therapeutic alliance. Future development of LLM-powered mental health support should continue to explore the balance between thorough assessment and immediate problem-solving needs, as well as using interviews in addition to self-reported ratings to collect higher-quality data.

**Acknowledgments**
This work is supported, in part, by the National Institutes of Health (NIH) Agreement No.1OT2OD032581 and R21NR020634, and the Rita and Alex Hillman Foundation Emergent Innovation Program.